\documentclass[10 pt, conference]{IEEEtran}
\usepackage[square,numbers]{natbib}
\usepackage{graphicx}
\graphicspath{{../imgs/}}
\usepackage{amsmath,amssymb,amsfonts}
\usepackage{algorithmic}
\usepackage{textcomp}
\usepackage{array}
\usepackage{float}
\usepackage{xcolor}
\usepackage{url}
\PassOptionsToPackage{bookmarks=false}{hyperref}

\def\BibTeX{{\rm B\kern-.05em{\sc i\kern-.025em b}\kern-.08em
    T\kern-.1667em\lower.7ex\hbox{E}\kern-.125emX}}

\begin{document}
\title{New Insights on Learning Rules for Hopfield Networks: Memory and Objective Function Minimisation}

\author{\IEEEauthorblockN{Pavel Tolmachev}
\IEEEauthorblockA{\textit{Electrical and Electronic Engineering Department,} \\
\textit{The University of Melbourne,}\\
Melbourne, Australia \\
0000-0001-6024-0135}
\and
\IEEEauthorblockN{Jonathan H. Manton}
\IEEEauthorblockA{\textit{Electrical and Electronic Engineering Department,} \\
\textit{The University of Melbourne,}\\
Melbourne, Australia \\
0000-0002-5628-8563}
}

\maketitle
\begin{abstract}
Hopfield neural networks are a possible basis for modelling associative memory in living organisms. 
After summarising previous studies in the field, we take a new look at learning rules, exhibiting them as descent-type algorithms for various cost functions. We also propose several new cost functions suitable for learning. 
We discuss the role of biases --- the external inputs --- in the learning process in Hopfield networks. Furthermore, we apply Newton's method for learning memories, and experimentally compare the performances of various learning rules. Finally, to add to the debate whether allowing connections of a neuron to itself enhances memory capacity, we numerically investigate the effects of self-coupling.
\end{abstract}


%
\IEEEpeerreviewmaketitle

\section{Introduction}
\IEEEPARstart{H}{opfield} networks were introduced almost four decades ago in an influential paper by Hopfield~\cite{hopfield1982neural}. They have received considerable attention because they exhibit properties of a content-addressable memory, and it has been hypothesized that biological associative memory might operate according to the same principles~\cite{pereira2018attractor, lansner2009associative, recanatesi2015neural}.

Hopfield network is a fully-connected directed graph with the vertices associated with neurons and edges corresponding to the strengths of interactions between these neurons, numerically characterised by \textit{weights}. In a discrete version of a Hopfield network the neurons can only be in one of two states: ``up'' and ``down''. The state of a given neuron depends on the influence of all the neighboring neurons and a \textit{bias}; the latter characterises a tendency of a neuron towards being in one of the states. 

The network evolves in discrete time-steps. At each time-step, a weighted sum of states of all the neurons and the neuron's bias (\textit{net input}) is computed at each neuron. The weights in the sum are the synaptic strengths afferent to a postsynaptic neuron. The neuron then re-evaluates its own state according to the following rule: if the resulting net input is positive, the neuron assumes ``up''-state, otherwise, it is updated with ``down''-state. Guided by this principle, the network is guaranteed to converge either to a stable configuration or a stable limit cycle after a finite number of time-steps. 

Discrete Hopfield network, although stripped from all the complexities associated with intrinsic neural dynamics has an intriguing property: manipulating the weights of the network one may able to imprint binary vectors (we refer to these vectors as \textit{patterns}) into the system. If the system is initialised sufficiently close to one of these memorised states (i.e. we present a distorted pattern to the network), the network recovers the original pattern (which corresponds to one of the stable configurations). The network may store more than just one pattern: the more patterns are written into the network, the less noise the network may tolerate while still correctly recovering a pattern. 

\subsection{Related work}
The capacity of the Hopfield Network to store patterns depends on the algorithm of tuning the weights between the neurons (\textit{learning rule}), so developing new learning strategies became a primal focus of researchers in this area. 
The first learning rule applied to store patterns was Hebbian learning \cite{hopfield1982neural}. To increase the storing capacity of the networks, various other learning rules were proposed \cite{storkey1999cient}, \cite{gardner1988space}, \cite{krauth1987learning}, \cite{personnaz1986collective}, \cite{diederich1987learning}, \cite{gardner1988space}. 
For the detailed review, a general family of rules was formulated and considered in \cite{abbott1990learning} and \cite{davey2004high}.

From purely engineering standpoint, in the recent literature Hopfield neural networks have been applied for image processing \cite{pajares2010hopfield}, \cite{hillar2014hopfield}, solving various combinatorial problems \cite{wen2009review} \cite{smith1998neural}, \cite{li2016modified}, random numbers generation \cite{tirdad2010hopfield}, \cite{hameed2018utilizing}, and have even been used in conjunction with Brain Computer Interfaces \cite{hsu2012application}, \cite{taghizadeh2013eeg}. 
To increase the energy efficiency of the implementation of Hopfield Networks in hardware, learning in sparse networks have been considered in \cite{tanaka2019spatially} and subsequently tested in the context of image restoration.

\subsection{Contributions}
In this paper we provide a new approach for generating learning rules by formulating a task of learning as an optimisation problem and applying descent-type strategies (Sec.\ref{sec: descent_type_rules}). Further, we show that some of the previously described rules could be obtained from the aforementioned perspective. In section \ref{sec: results} we demonstrate the performance of the newly proposed learning rules and made a comparison with the rules previously described in the literature. We then extend these learning rules with the application of Newton's method. It was debated weather the neural self-connections benefit the memory capacity of the Hopfield Networks \cite{davey2004high} \cite{gorodnichy1999optimal} \cite{kanter1987associative}. We experimentally studied the effects of self-coupling in \ref{sec: self_coupling}. 
Along the way, we discuss the role the biases play in learning in the Hopfield Networks (Sec. \ref{sec: tuning_biases}). 

\section{Definition of the Model}
In a concise form, the model could be summarised as follows:

\begin{equation}
\begin{split}
& h_i[n] = \sum_{j=1}^N \omega_{ij} s_j[n-1] + b_i \\
& s_i[n] = \text{sign}(h_i[n]), \quad i \in \{1, 2 \dots, N\} 
\end{split}
\label{eq: dynamics}
\end{equation}
Where $h_i$ - net input to a neuron $i$ (also could be referred to as hidden variable or local field), $\omega_{ij}$ - the synaptic strength from a neuron $i \to j$, $s_i$ - the state of a neuron $i$, $b_i$ - bias of a neuron. Number $n$ in square brackets denotes the number of the current time-step. 

In a vectorised from, the model's dynamic could be stated:

\begin{equation}
\begin{split}
& \boldsymbol{h}[n] = \boldsymbol{W s}[n-1] + \boldsymbol{b} \\
& \boldsymbol{s}[n] = \text{sign}(\boldsymbol{h}[n])
\end{split}
\label{eq: dynamics_vect}
\end{equation}

where $\boldsymbol{W}$ is a matrix containing $\omega_{ij}$ as elements.

\subsection{An example of a learning rule: Hebbian learning}
An obvious candidate for learning procedure would be the Hebbian rule, which was stated as \textit{"cells which fire together wire together"} back in 1948 by Donald Hebb \cite{hebb2005organization}. The intuition behind this rule in the context of the Hopfield networks is simple: suppose the system initially has all the weights set to zero, and one wants to write some pattern into a system.
The obvious thing to do would be to increase the weights between those neurons which fire together by a constant value, to facilitate their parallel alignment. Conversely, if the neurons are aligned antiparallely in the pattern to memorise, the weights should be decreased by the same amount.
To put succinctly, the Hebbian rule is defined:
\begin{equation}
\Delta w_{ij} = \lambda \sigma_{i} \sigma_{j}
\label{eq: hebbian_learning_rule (1)}
\end{equation}
Where $\sigma_i$ is the state of a neuron in the pattern to be memorised. $\lambda$ is the proportionality coefficient. 
If there are multiple patterns, the rule then becomes:
\begin{equation}
\Delta w_{ij} = \lambda \sum_{\nu = 1}^p \sigma_{i}^{\nu} \sigma_{j}^{\nu}
\label{eq: hebbian_learning_rule (2)}
\end{equation}

Where index $\mu \in {1, \dots, p}$ denotes the number of a pattern to memorise, and $p$ is the total number of patterns.

\section{Important aspects of learning in the Hopfield networks} \label{sec: important_aspects}

\subsection{Physical analogy}
If one considers only symmetric weight matrices, and thus, symmetric interactions between the neurons, there is a useful analogy from physics: an evolution of a system of atoms with magnetic moments interacting with one another could also be described by the same dynamics.
The physical analogy allows us to consider the evolution of a Hopfield Network as a relaxation of a physical system into a state achieving a local minimum of energy $E$. 

\begin{equation}\label{eq: energy}
    E = - \frac{1}{2}\boldsymbol{s}^T \boldsymbol{W} \boldsymbol{s} - (\boldsymbol{b}, \boldsymbol{s})
\end{equation}

Presenting the system with the initial state, from which it then continues to evolve, the system relaxes to a closest local minimum. These minima are defined by the weight matrix. Thus, the task of memorising the patterns could then be seen as creating funnels in the energy landscape: the rounder and deeper the funnels are, the more robustly the pattern could be recovered.
During the process of learning the patterns, the system may also create spurious minima in the energy landscape, which do not correspond to any of the patterns it has learned. The trivial examples of spurious minima would be anti-patterns.

\subsection{Biological plausibility}
The important aspect of a learning rule is its biological plausibility, as the Hopfield Networks are studied mostly to understand how the biological associative memory works. Here, by using term "biologically plausible" in relation to a learning rule we do not claim that the learning in biological systems occurs according to the rule. Rather, we use the term to eliminate non-plausible learning schemas.

The two main components defining biological plausibility are \textit{locality} and \textit{incrementality}. 
For the learning rule to be \textit{strictly local}, the update of weights has to rely only on the information available at a given synapse. The rule could also be said to be \textit{loosely local} if the update requires information from the other synapses from presynaptic and/or postsynaptic neurons.

Incrementality of a learning rule states that updates of the weights may depend only on one pattern at the time. Incremental learning is closely linked to forgetting. As the network one patter after another, it eventually overrides the information about the old patterns encrypted in its weights \cite{nadal1986networks}.

One can see that Hebbian learning falls into the category of biologically plausible rules in the aforementioned sense, as it is both local and incremental.

\subsection{Asymmetry in the connectivity}
Although it is much easier to operate with the symmetric weight matrices, as it provides a useful analogy with spin glasses \cite{hopfield1982neural}, the biological networks do not exhibit such property. The asymmetry in the connectivity is still largely unexplored phenomena, despite the bulk of research done in this area. It has been claimed that asymmetry in the synaptic weights allows storing limit cycles, and thus, oscillations \cite{hopfield1982neural}. Further pursuing this line of research may shed some light on how to construct Central Pattern Generators (CPGs). 
In a letter \cite{parisi1986asymmetric}, the author noted that if the process of updating the weights never stops in the real network, the sufficient asymmetry of weights might be needed for the network to perform successfully. Otherwise, the system may stuck in a spurious minimum and then reinforce it through continuous learning. To counter this, if the system has not reached a stable pattern, it has to relapse into a chaotic behaviour, for the weights not to be excessively updated in some particular direction.

Although the exploration of the role of asymmetry in these networks is tempting, in this paper, we consider only systems with the symmetric weights.

\subsection{Linear programming and learning in Discrete Hopfield Networks} \label{sec: linearity_of_learning}
During the learning stage, a pattern to be memorised is clamped to a network, effectively forcing the neurons to be in a predefined state: $s_i = \sigma_i \quad \forall i \in \{1, 2, \dots, N\}$. Furthermore, since the states of neurons are fixed, a net input on the neuron $i$ depends only on the clamped pattern $\boldsymbol{\sigma}$ and weights presynaptic to the neuron, $\boldsymbol{\omega_i}$. For the pattern to be stable, each neuron has to adjust its net input to have the same sign as the state-value clamped to it. To achieve this, a neuron needs to tune only the afferent weights, while it does not require any information about the synaptic strengths between any other neurons. 

Let's define $h_i^{\mu}(\boldsymbol{\omega_i}) = \sum_{j=1}^{N} \omega_{ij} \sigma_j^{\mu}$ to be a net input to a neuron $i$ with the pattern $\mu$ clamped on the network. Then, for a given pattern $\boldsymbol{\sigma^{\mu}}$, one can formulate a constraint to be imposed on the weights afferent to a neuron $i$:
\begin{equation}
h_i^{\mu}(\boldsymbol{\omega_i}) \sigma_i^{\mu} \geq 0, \quad \mu = \{1, \dots, p\}
\end{equation}

Or, in a canonical LP form ($a^T x + b \leq 0$):
\begin{equation}
 -\sigma_i^{\mu}\boldsymbol{(\sigma^{\mu})}^T \boldsymbol{\omega_i} - \sigma_i^{\mu} b_i \leq 0, \quad \mu \in \{1, \dots, p\}
\end{equation} \label{eq: constraints}

Further down, we are going to address these constraints as \textit{stability constraints}. Note that these constraints depend on the weights and a bias linearly. Because of this, the task of learning breaks down into a set of linear programs, each defined for a particular site $i$, and solved independently from the other neurons. We refer to such property of Discrete Hopfield Network as \textit{linearity of learning}.

Developing the idea of linearity of learning further, one can obtain a useful geometric perspective on the learning algorithms. The stability constraints correspond to hyper-planes in weights-biases space, which pass through the origin, and each of the constraints defines a feasible half-space. Thus, if the solution to the learning problem exists, it lies in the pointed multi-dimensional cone defined by the intersection of the half-spaces. To make the solution robust, one needs to stay as far from the boundaries defined by the stability constraints as possible. These will be ensured if the weights and a bias lie on the bisectrix of the feasible cone in the weights-bias space.


To counter the unbounded growth of the weights, either regularisation or additional constraint on the weight's norm has to be imposed.


\subsection{Weights re-scaling} \label{sec: rescaling}
Another useful concept may be obtained by considering the alignment of a net input $h_i \sigma_i$ on a neuron $i$. It is easy to notice, that the stability constraints are invariant to the scaling of the incoming weights and a bias of a neuron by some positive factor. Thus, if the local linear program on a neuron $i$ is successfully solved, and a pair ($\boldsymbol{\omega_i}, b_i)$ is one of the solutions, a pair $(a\boldsymbol{\omega_i}, ab_i)$ also satisfies the stability constraints, for any $a > 0$ 

\section{Posing the task of learning as an optimisation problem} \label{sec: descent_type_rules}
The problem of robust pattern retrieval is conjugated with making the basins of attractions of the patterns as wide and round as possible.  However, it is highly nontrivial to characterise the width and roundness of basins of attraction analytically, although some experimental investigations have suggested that maximising the quantity $\gamma =  \frac{h_i \sigma_i}{\| \boldsymbol{\omega_i}\|_2}$ increases the size of basins of attraction and prevents the unbound growth of elements of $\boldsymbol{\omega_i}$ \cite{kepler1988domains}.

Instead of employing a direct approach, we replace the task of learning a pattern by the task of making the system to maintain the pattern, once it has reached this state. 

That being said, the new task can be expressed as the following optimisation problem:
\begin{equation}
\min_{W} D \big(\boldsymbol{\sigma}, \text{sign}(\boldsymbol{W\sigma} + \boldsymbol{b})\big)
\end{equation}
where $D$ is some distance measure between two binary vectors. The first variable is one of the learned patterns which the network has started from, and the second argument is the state of the system after the one round of evolution started from this pattern.

This problem, in turn, may be further simplified down to:
\begin{equation}
\min_{W} D \big(\boldsymbol{\sigma}, \lambda(\boldsymbol{W\sigma} + \boldsymbol{b})\big)
\end{equation}
Where we simply omitted the $sign$ function and replaced with identity function multiplied by some positive $\lambda$. 
\subsection{Learning as a gradient descent}
The reformulation of a problem allows us to implicitly calculate the update of weights and biases, for the whole function to be minimised by setting the change of the learned parameters to be proportional to the anti-gradient of the function.

\subsubsection{Descent Overlap}
Utilising the negative cosine similarity measure instead of a distance function:
$D \big(\boldsymbol{\sigma}, \lambda(\boldsymbol{W\sigma} + \boldsymbol{b})\big) = -\big(\boldsymbol{\sigma}, \lambda(\boldsymbol{W\sigma} + \boldsymbol{b)}\big)$.
One arrives at the following update rule:
\begin{equation}
\begin{split}
& \Delta \omega_{ij} = \lambda \sigma_i \sigma_j \\
& \Delta b_i = \lambda \sigma_i
\end{split}
\end{equation}
Which is exactly the Hebbian rule described previously (see eq. \ref{eq: hebbian_learning_rule (1)}). It is also worth noticing that minimising the negative cosine similarity measure corresponds to minimising an energy function (eq. \ref{eq: energy}).

\subsubsection{Descent L2 norm}
If one uses the L2 norm for the distance measure $D$ between the vectors, the resulting update rule will take the form:
\begin{equation}
\begin{split}
& \Delta \omega_{ij} = 2\lambda \big(\sigma_i - \lambda h_i\big)\sigma_j \\
& \Delta b_i = 2\lambda \big(\sigma_i - \lambda h_i\big)
\end{split}
\end{equation}
The rule is aimed to minimise the same function as the pseudo-inverse learning rule, but at the same, it is biologically plausible in a defined above sense, as it is local and incremental. Interesting enough, apart from the factor $2$, it is actually the "rule II" in Diederich and Opper \cite{diederich1987learning} but obtained from a different perspective.

This rule is also could be seen from yet another angle: $(\sigma_i - \lambda h_i)^2 = 1 - 2\lambda h_i \sigma_i + \lambda^2 h_i^2$, so it follows that this rule seeks to maximise the alignment of the net input with the clamped pattern (term $- 2\lambda h_i \sigma_i$), while preventing the local-field from growing too much (regularisation $\lambda^2 h_i^2$). 

\subsubsection{Descent L1 norm}
Using L1 distance as $D$ (which is equivalent to minimising the Hamming distance) one gets:
\begin{equation}
\begin{split}
& \Delta \omega_{ij} = \lambda \, \text{sign} \big(\sigma_i - \lambda h_i\big)\sigma_j \\
& \Delta b_i = \lambda \, \text{sign} \big(\sigma_i - \lambda h_i\big)
\end{split}
\end{equation}
This rule is, in fact, exactly the same as the 'rule I' proposed in Diederich and Opper \cite{diederich1987learning}:
$$ \text{sign} \big(\sigma_i - \lambda h_i\big) = \text{sign} \big(\sigma_i (1 - \lambda h_i \sigma_i)\big) = \sigma_i \, \text{sign} \big(1 - \lambda h_i \sigma_i\big) $$



\subsection{Robust learning with barrier functions}
For $p$ patterns to be learned, each particular neuron has a set of $p$ constraints, where each constraint defines an allowed half-space. To counteract the noise, the weights of incoming connections to a neuron have to stay away from the hyper-planes corresponding to constraints as far as possible. To achieve this, we propose an exponential barrier $\exp\{-\lambda \sigma_i^{\mu}h_i^{\mu}\}$ which grows fast if the constraints are not satisfied. Other choices of the barrier function are possible. The tendency of a system to stay away from the constraints will lead the weights and biases to become infinitely large. To deal with this issue, one has to impose a regularisation on learning variables to keep the norms of weights and biases from infinite growing. It should be noted, that if the problem is infeasible, minimisation of this function will produce a parameter set which tends to minimise the number of broken constraints.
The resulting minimisation problem (for one neuron) is formulated as:
\begin{equation}
\min_{\omega_i} \sum_{\mu = 1}^p e^{-\lambda \sigma_i^{\mu}\big((\boldsymbol{\sigma^{\mu}}, \boldsymbol{\omega_i}) + b_i\big)} + \frac{\alpha}{2} (\|\omega_i\|_2^2 + b_i^2)
\end{equation}

Since this function is readily differentiable, one can obtain an update rule as follows:
\begin{equation}
\begin{split}
& \Delta \omega_{ij} = \lambda \sum_{\mu = 1}^p \sigma_i^{\mu} \sigma_j^{\mu} e^{-\lambda \sigma_i^{\mu} h_i} - \alpha\omega_{ij} \\
& \Delta b_i = \lambda \sum_{\mu = 1}^p \sigma_i^{\mu} e^{-\lambda \sigma_i^{\mu} h_i}  - \alpha b_i 
\end{split}
\end{equation}
Of course, this rule is not biologically plausible in the previously defined sense, as it requires access to the information about all the patterns at once. 
Nevertheless, one can easily construct a biologically plausible version of it just by considering one constraint at the time (and, hence, just one pattern). Thus, the learning automatically incorporates forgetting, is local and incremental.

\subsection{Tuning the biases} \label{sec: tuning_biases}
If one employs the approach of learning through gradient descent, it becomes obvious that the biases also may contribute to the process of memorising (although they were largely neglected in previous research papers, being dismissively set to zero). The biases could be regarded as the weights of connections coming from an up-stated neuron from outside of the network. From now on, to use a common notation for weights and biases, we will use $\boldsymbol{\omega'_i}$ to refer to a vector of $\boldsymbol{\omega_i}$ and $b_i$ concatenated together, and the symbol $\boldsymbol{\sigma'^{\mu}}$ will denote the original pattern $\boldsymbol{\sigma^{\mu}}$ concatenated with $1$.

\subsection{Beyond the gradient descent: Newton's method}
To speed up the convergence of learning algorithm, one may consider the learning with variable step-size, instead of fixed $\lambda$.
To obtain the step-size as a function of locally available information, one has to find an explicit expression for the inverse of the Hessian of an objective function.

As an example, let us consider the task of minimisation of the function:

\begin{equation}
f(\boldsymbol{\omega'_i}) = \frac{1}{2}\sum_{\mu = 1}^p(\lambda h_i^{\mu} - \sigma_i^{\mu})^2 + \frac{\alpha}{2}\|\boldsymbol{\omega'_i}\|_2^2 
\end{equation}

Then the gradient of this function is written as:

\begin{equation}
\boldsymbol{\frac{\partial f}{\partial \omega'_{i}}} = \lambda \sum_{\mu = 1}^p(\lambda h_i^{\mu} - \sigma_i^{\mu}) \boldsymbol{\sigma'^{\mu}} + \alpha \boldsymbol{\omega'_i}
\label{eq: gradient}
\end{equation}

And the Hessian:

\begin{equation}
\boldsymbol{H_{\omega'_i}} = \boldsymbol{\frac{\partial^2 f}{\partial \omega_{i}^{'2}}} = \alpha \boldsymbol{I} + \lambda^2  \boldsymbol{Z'} \boldsymbol{Z'}^T
\end{equation}

where $\boldsymbol{Z'}$ is the matrix, with the augmented patterns $\boldsymbol{\sigma'^{\mu}}$ as columns.
And the learning rule would be:
\begin{equation}
\Delta \boldsymbol{\omega'_{i}} = -\boldsymbol{H_{\omega'_i}}^{-1} \boldsymbol{\frac{\partial f}{\partial \omega'_{i}}}
\label{eq: newtons_update}
\end{equation}

Luckily, Hessian $\boldsymbol{H_{\omega'_i}}$ is easily invertible thanks to Neumann series for matrices. Applying $(\boldsymbol{I} - \boldsymbol{A})^{-1} = \sum_{k=0}^{\infty} \boldsymbol{A}^k$:
\begin{equation}
\boldsymbol{H_{\omega'_i}}^{-1} =  \frac{1}{\alpha} (\boldsymbol{I} - \lambda \boldsymbol{Z'}\boldsymbol{Z'^{T}} + \lambda^2 \boldsymbol{Z'}\boldsymbol{C'} \boldsymbol{Z'}^{T} - \lambda^3 \boldsymbol{Z'}\boldsymbol{C'}^{2} \boldsymbol{Z'}^{T} + \dots)
\end{equation}

where $\boldsymbol{C'} = \boldsymbol{Z'}^{T} \boldsymbol{Z'}$ - the correlation matrix.
Considering learning all the patterns at once, one may use only the first two terms to get a reasonable approximation.
If the task is to learn sequentially, just one pattern at the time, the inverse of the Hessian may be computed explicitly:
\begin{equation}
\label{eq: Hessian_incremental}
\boldsymbol{H_{\omega'_i}}^{-1} =  \frac{1}{\alpha} (\boldsymbol{I} -  \frac{\lambda}{1 + \lambda (N + 1)}\boldsymbol{\sigma'}\boldsymbol{\sigma'}^T)
\end{equation}

with $\lambda (N + 1) < 1$ for the infinite series to converge.

We have arrived at a modification of previously described set of learning rules: a weight's update (\ref{eq: newtons_update}) with the gradient (\ref{eq: gradient}) and an inverse of a Hessian in (\ref{eq: Hessian_incremental}). 

\section{Methods} \label{sec: methods}
This section describes numerical experiments preformed in the Results section.

We will refer to the rules described in section \ref{sec: descent_type_rules} as \textit{descent-type rules}. They are summarised in the table below (Table \ref{tbl: descent_type_rules}).

\begin{table}[!h]
\renewcommand{\arraystretch}{2.2}
\caption{Descent-type rules}
\label{tbl: descent_type_rules}
\centering
\begin{tabular}{|c||c|}
\hline
$\textbf{Learning rule}$  & $\textbf{Minimised Function}$\\
\hline
DescentL2 & $\displaystyle \sum_{\mu = 1}^{p}(\lambda h_i^{\mu} - \sigma_i^{\mu})^2$ \\
\hline
DescentL1 & $\displaystyle  \sum_{\mu = 1}^{p}|\lambda h_i^{\mu} - \sigma_i^{\mu}|$ \\
\hline
DescentExpBarrier & $\displaystyle  \sum_{\mu = 1}^{p} exp\big(-\lambda h_i^{\mu} \sigma_i^{\mu}\big)$ \\
\hline
DescentExpBarrierSI* & $\displaystyle  \sum_{\mu = 1}^{p} exp\Big(-\frac{\lambda h_i^{\mu} \sigma_i^{\mu}}{\|\sum_j \omega_{ij}^2 + b_i\|}\Big)$ \\
\hline
\end{tabular}
\\
*postfix "SI" stands for scale-invariant (see sections \ref{sec: rescaling}, \ref{sec: descent_type_rules})
\end{table}

Since employing Newton's method does not affect the performance in terms of accuracy of recall (it only speeds up the convergence), we have used information from a Hessian whenever it was possible. Except for DescentExpBarrierSI rule, all the descent-type rules were combined with the L2 regularisation on weights and biases.
We will compare incremental rules separately from the non-incremental ones. For our simulations, a network of 75 neurons was used. All the tests were performed with random patterns, where the probability of one particular neuron to be ``on'' was 0.5. For each of the learning rules, we have made the network to learn $p$ patterns, where $p$ varies from 1 to 75. To test the recall performance, for each number $p$, we have sequentially introduced $k$ number of random flips (changes of a neuron's state to an opposite one) into one of the patterns, and presented the distorted pattern to the network ($k$ is running from 1 to 37). After each retrieval, the weights and the biases of the network were reset to some small random values drawn from a gaussian distribution with a zero mean. The whole cycle was repeated for 100 times. Thus, the total number of retrievals was 100 $\times$ 75 $\times$ 37 for each rule. For each retrieval, the overlap (an inner product divided by $N$) between the retrieved vector and the intended pattern was computed. The visual measure we have used to evaluate the performance of learning prescriptions was the position of the curve $p_{\epsilon}(k)$, on which the overlap value was $\epsilon = 0.95$ so that the overlap values below the curve were higher than $\epsilon$: the greater the area under the curve, the better the performance of a given rule. All the code is publicly available on GitHub: \texttt{\url{github.com/ptolmachev/Hopfield_Nets}}

\section{Results and Discussion} \label{sec: results}
In this section, we make a comparison of newly proposed learning strategies with some of the previously known learning rules.  

We have made a comparison of the following incremental learning rules: Hebbian, Storkey, Diederich and Opper rules I and II and all of the descent-type learning rules (Table \ref{tbl: descent_type_rules}). We also have made comparison with the rule proposed by Krauth and Mezard \cite{krauth1987learning}, since it could be attributed neither to incremental, nor to non-incremental rules since this rule employs weak-pattern-first update strategy. We also have used a modification of Gardner's rule \cite{gardner1988space}: combining it with Krauth-Mezard strategy of choosing the next pattern to learn (we refer to it as Gardner-Krauth-Mezard learning rule).

\begin{table}[!t]
\renewcommand{\arraystretch}{1.5}
\caption{Incremental rules and parameters}
\label{tbl: table_incremental_rules}
\centering
\begin{tabular}{|c||c|}
\hline
$\textbf{Learning rule}$  & $\textbf{Parameters}$\\
\hline
Hebbian & $\text{sc} = True$, $\lambda = 1$\\
\hline
Storkey & $\text{sc} = True$, $\lambda = 1$\\
\hline
Diederich-Opper I & $\text{sc} = True$, $\text{lr} = 10^{-2}$\\
\hline
Diederich-Opper II & $\text{sc} = True$, $\text{lr} = 10^{-2}$, $\text{tol} = 0.1$\\
\hline
Krauth-Mezard & $\text{sc} = True$, $\text{lr} = 10^{-2}$, $\text{maxiter} = 200$\\
\hline
DescentExpBarrier & $\text{sc} = True$, $\text{tol} = 0.1$, $\lambda = 0.5$, $\alpha = 10^{-3}$\\
\hline
DescentL1 & $\text{sc} = True$, $\text{tol} = 0.1$, $\lambda = 0.5$, $\alpha = 10^{-3}$\\
\hline
DescentL2 & $\text{sc} = True$, $\text{tol} = 0.1$, $\lambda = 0.5$, $\alpha = 10^{-3}$\\
\hline
DescentExpBarrierSI & $\text{sc} = True$, $\text{tol} = 0.1$, $\lambda = 0.5$\\
\hline
Gardner-Krauth-Mezard & $\text{sc} = True$, $\text{lr} = 10^{-2}$, $k = 1.0$, \\
 &  $\text{maxiter} = 100$ \\
\hline
\end{tabular}
\\ 
*$sc$ - self-coupling, $lr$ - learning rate, $tol$ - criteria for optimisation termination, $maxiter$ - maximum length of sequence of patterns to learn
\end{table}

For non-incremental rules comparison, we have chosen Hebbian, pseudo-inverse \cite{personnaz1986collective}, Storkey \cite{storkey1997increasing}, all of the descent-type learning rules, and a Krauth-Mezard rule \cite{krauth1987learning}.

\begin{table}[!t]
\renewcommand{\arraystretch}{1.5}
\caption{Non-incremental rules and parameters}
\label{tbl: table_nonincremental_rules}
\centering
\begin{tabular}{|c||c|}
\hline
$\textbf{Learning rule}$  & $\textbf{Parameters}$\\
\hline
Hebbian & $\text{sc} = True$, $\lambda = 1$ \\
\hline
Storkey & $\text{sc} = True$, $\lambda = 1$\\
\hline
Pseudoinverse & ---\\
\hline
Krauth-Mezard & $\text{sc} = True$, $\text{lr} = 10^{-2}$, $\text{maxiter} = 200$\\
\hline
DescentExpBarrier & $\text{sc} = True$, $\text{tol} = 10^{-3}$, $\lambda = 0.5$, $\alpha = 10^{-3}$\\
\hline
DescentL1 & $\text{sc} = True$, $\text{tol} = 10^{-3}$, $\lambda = 0.5$, $\alpha = 10^{-3}$\\
\hline
DescentL2 & $\text{sc} = True$, $\text{tol} = 10^{-3}$, $\lambda = 0.5$, $\alpha = 10^{-3}$\\
\hline
DescentExpBarrierSI & $\text{sc} = True$, $\text{tol} = 10^{-3}$, $\lambda = 0.5$\\
\hline
Gardner-Krauth-Mezard & $\text{sc} = True$, $\text{lr} = 10^{-2}$, $k = 1.0$, \\
 &  $\text{maxiter} = 100$ \\
\hline
\end{tabular}
\\ 
*$sc$ - self-coupling, $lr$ - learning rate, $tol$ - criteria for optimisation termination, $maxiter$ - maximum length of sequence of patterns to learn
\end{table}

\subsection{Comparison of incremental rules}
\begin{figure}[!t]
\centering
\includegraphics[width=0.45\textwidth]{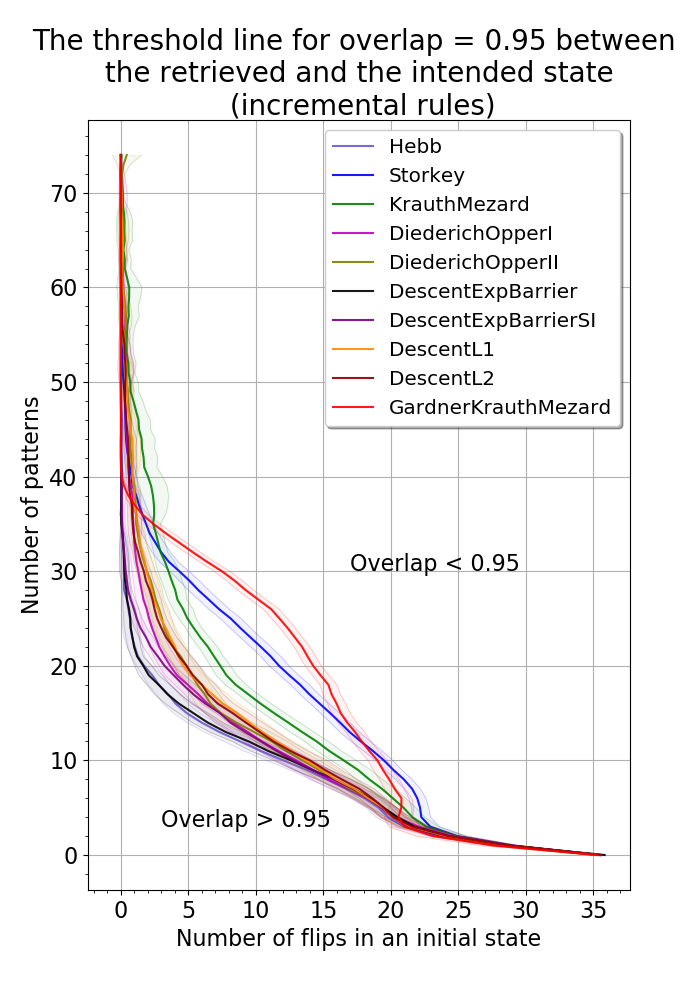}
\caption{The $p_{\epsilon}(k)$ curve for incremental rules $\epsilon = 0.95$, $N = 75$ (description in Sec. \ref{sec: methods}). The curve denotes the threshold at which the normalised dot product between recovered and intended state is 0.95. The greater the area under the curve the better the performance of a learning rule.}
\label{fig: incremental_rules_comparison}
\end{figure}

The performance of incremental class rules is depicted in figure \ref{fig: incremental_rules_comparison}. Not surprisingly, Gardner-Krauth-Mezard rule performs better than the other rules since the rule relies on weak-pattern-first update strategy, although simpler Krauth-Mezard rule outperforms other rules in the high load regime. Storkey rule is the rule with the best storing capacity if we consider only sequential updates in a predefined order. From theoretical considerations, it was also anticipated that DescentL1 and DescentL2 would perform analogously to Diederich and Opper rules I and II correspondingly \cite{diederich1987learning}.

\subsection{Comparison of non-incremental rules}
\begin{figure}[!t]
\centering
\includegraphics[width=0.45\textwidth]{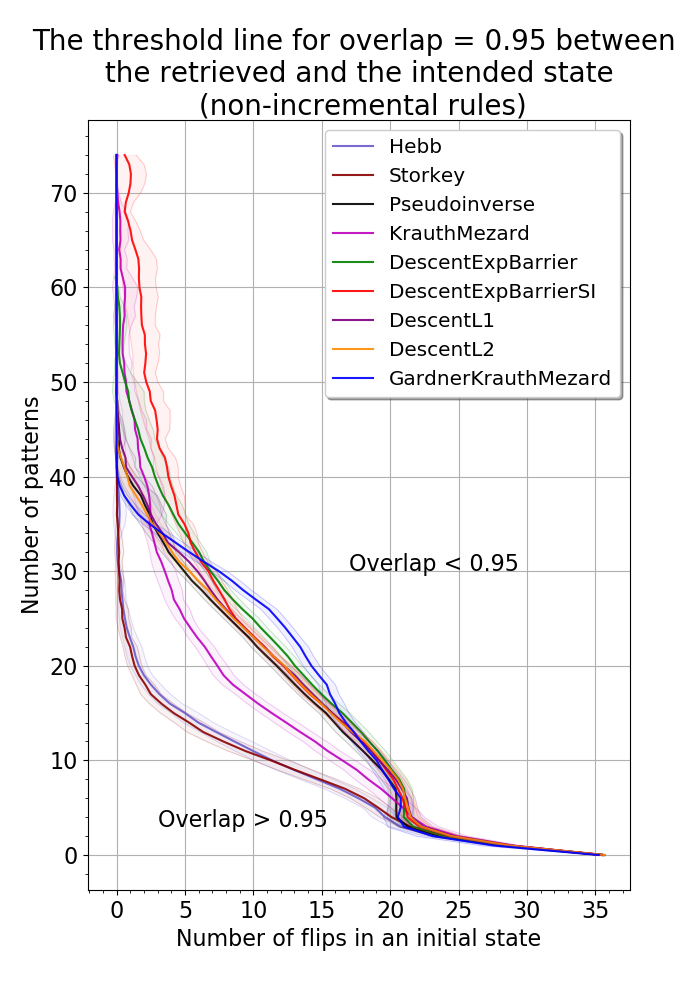}
\caption{The $p_{\epsilon}(k)$ curve for non-incremental rules $\epsilon = 0.95$, $N = 75$ (description in Sec.  \ref{sec: methods}). The curve denotes the threshold at which the normalised dot product between recovered and intended state is 0.95. The greater the area under the curve the better the performance of a learning rule.}
\label{fig: nonincremental_rules_comparison}
\end{figure}

The performance is depicted in Figure \ref{fig: nonincremental_rules_comparison}.
The storing capacity of Hopfield Network with Gardner-Krauth-Mezard rule is still comparable to the ones of best non-incremental rules: it outperforms any other non-incremental rule in the moderate-flips-moderate-patterns region. In the high-load region, the newly proposed DescentExpBarrierSI exhibits higher memory capacity than any other rule.
As could be seen in the figure, we have experimentally confirmed that the performance of a DescentL2 rule is similar to the one of a pseudo-inverse rule, since these rules minimise the same quantity. There is also not much difference in performance between the DescentL1 rule and the pseudo-inverse, as the corresponding curves are practically aligned. Krauth-Mezard rule handles memorising lots of patterns quite well (the upper-left region in the figure), but it fails to keep up with the descent-type rules when there are fewer memories needed to be retained. Finally, although one of the best incremental rules, Storkey rule performs on par with the Hebbian prescription, when the update is done non-incrementally. This could be anticipated, since the net inputs $h_i$ are close to zero when the network is in a blank-slate state, and one-off non-incremental update effectively renders the rule to be an instance of Hebbian learning.

\subsection{Dependence of the performance on self-connectivity} \label{sec: self_coupling}
In this section, we address the following question: how self-connectivity affects learning performance? It has been debated whether self-coupling is beneficial for memory storage in \cite{davey2004high} \cite{kanter1987associative} \cite{gorodnichy1999optimal}, but, it seems, the issue is still not resolved. Here, we give experimental results to investigate the effects of self-coupling on the performance of various learning rules, namely, Hebbian, DescentL2 (Pseudo-inverse), Gardner-Krauth-Mezard and DescentExpBarrierSI. We present the comparison only for non-incremental learning. The parameters for the simulations, apart from self-coupling parameter, are the same as presented in the table \ref{tbl: descent_type_rules}. 

\begin{figure}[!h]
\centering
\includegraphics[width=0.45\textwidth]{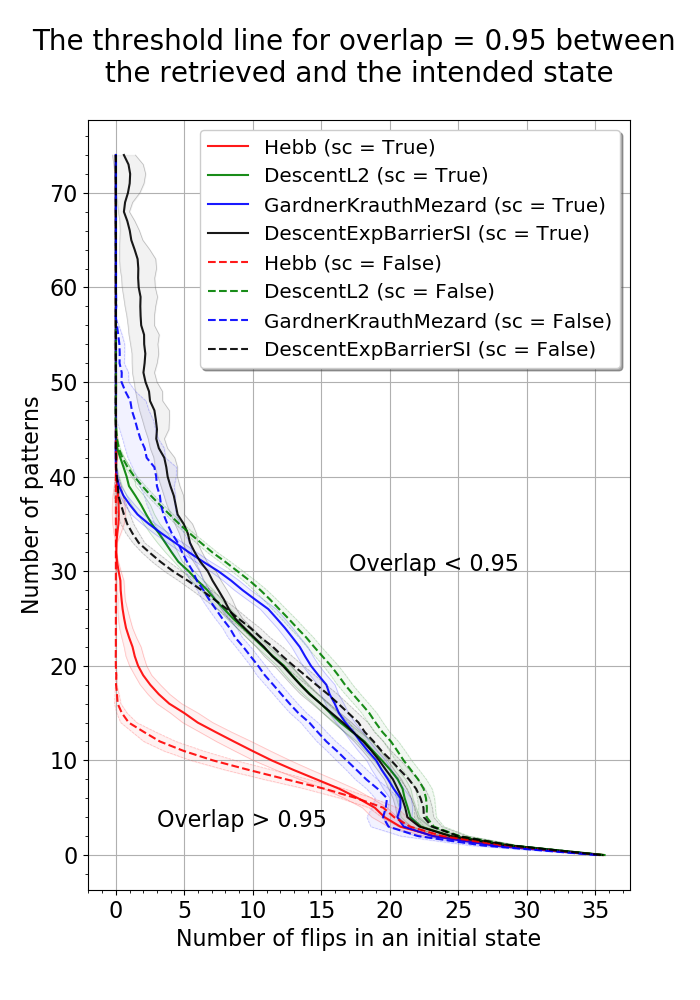}
\caption{The $p_{\epsilon}(k)$ curve for non-incremental rules $\epsilon = 0.95$, $N = 75$ with varying self-connectivity parameter. The curve denotes the threshold at which the normalised dot product between recovered and intended state is 0.95. The greater the area under the curve the better the performance of a learning rule.}
\label{fig: sc_comparison}
\end{figure}

From the results depicted in Fig. \ref{fig: sc_comparison}, it could be inferred that self-coupling may be preferable for some learning prescriptions. For instance, the performance of the Hebbian learning only gets better with the self-coupling parameter turned on. The positive effect of self-connectivity is also apparent for the Gardner-Krauth-Mezard learning rule in the high noise region. DescentExpBarrierSI demonstrate superior performance in high-load-low-noise region when the self-connectivity is present. In agreement with the investigations done in \cite{gorodnichy1999optimal}, DescentL2 (and its one-off equivalent pseudo-inverse) learning rule has better performance when there is no self-connectivity. 

\subsection{Incremental learning vs. one-off update}
If one considers strictly incremental versus non-incremental learning, there is a trade-off between learning in portions and the robustness. When the learning is done in an incremental way, each time a new linear constraint is presented, the system has to adjust the parameters to satisfy it. As a result, the point in the weights-biases space, representing the weights of the network, is bouncing off from the hyper-plane corresponding to a newly imposed constraint. However, it may leap too far, so that some of the constraints set previously may be broken. If it is the case, the network forgets the pattern associated with the broken constraint.

As the online learning in living organisms is done incrementally, the forgetting is natural. Nevertheless, to counteract this problem species have likely developed some countermeasures to make the learning not so purely incremental. Several such mechanisms may be at work, including unlearning spurious states \cite{hopfield1983unlearning} \cite{kleinfeld1987unlearning} \cite{van1990increasing} and re-iterating amongst the weakest patterns (as have been proposed by Krauth and Mezard \cite{krauth1987learning}). Since these mechanisms could not be performed online, as that would interfere with the ongoing activities, it has been hypothesised that non-incremental learning may occur during the REM sleep \cite{crick1983function}.

\section{Conclusion}

We have presented new insights into how the learning may be performed in Hopfield Neural Networks using gradient-descent: we have relaxed the task of learning down to a problem of minimisation a smooth differentiable function. Using this principle, we have established the equivalence of some of the previously proposed learning algorithms to minimisation of the particular objectives. Guided by the described idea, we were able to formulate a new family of learning rules. We also have made an extensive comparison of several learning rules with the new ones we proposed in this paper. We have clarified the role of biases in the learning task: the biases could be regarded as weights of connections incoming from a neuron from outside of the network fixed in ``up''-state. The effects of the presence of self-connections on the memory capacity were simulated. Finally, we have discussed a trade-off between incremental and robust learning.


%



\section*{Acknowledgment}
Pavel Tolmachev would like to thank Andrei Pavlov for the invaluable discussions.

\ifCLASSOPTIONcaptionsoff
  \newpage
\fi



%
\newpage

\bibliographystyle{apalike}
\bibliography{references} 

\begin{thebibliography}{}

\bibitem[Abbott, 1990]{abbott1990learning}
Abbott, L.~F. (1990).
\newblock Learning in neural network memories.
\newblock {\em Network: Computation in neural systems}, 1(1):105--122.

\bibitem[Crick et~al., 1983]{crick1983function}
Crick, F., Mitchison, G., et~al. (1983).
\newblock The function of dream sleep.
\newblock {\em Nature}, 304(5922):111--114.

\bibitem[Davey et~al., 2004]{davey2004high}
Davey, N., Hunt, S.~P., and Adams, R. (2004).
\newblock High capacity recurrent associative memories.
\newblock {\em Neurocomputing}, 62:459--491.

\bibitem[Diederich and Opper, 1987]{diederich1987learning}
Diederich, S. and Opper, M. (1987).
\newblock Learning of correlated patterns in spin-glass networks by local
  learning rules.
\newblock {\em Physical review letters}, 58(9):949.

\bibitem[Gardner, 1988]{gardner1988space}
Gardner, E. (1988).
\newblock The space of interactions in neural network models.
\newblock {\em Journal of physics A: Mathematical and general}, 21(1):257.

\bibitem[Gorodnichy, 1999]{gorodnichy1999optimal}
Gorodnichy, D.~O. (1999).
\newblock The optimal value of self-connection.
\newblock In {\em IJCNN'99. International Joint Conference on Neural Networks.
  Proceedings (Cat. No. 99CH36339)}, volume~1, pages 663--668. IEEE.

\bibitem[Hameed and Ali, 2018]{hameed2018utilizing}
Hameed, S.~M. and Ali, L. M.~M. (2018).
\newblock Utilizing hopfield neural network for pseudo-random number generator.
\newblock In {\em 2018 IEEE/ACS 15th International Conference on Computer
  Systems and Applications (AICCSA)}, pages 1--5. IEEE.

\bibitem[Hebb, 2005]{hebb2005organization}
Hebb, D.~O. (2005).
\newblock {\em The organization of behavior: A neuropsychological theory}.
\newblock Psychology Press.

\bibitem[Hillar et~al., 2014]{hillar2014hopfield}
Hillar, C., Mehta, R., and Koepsell, K. (2014).
\newblock A hopfield recurrent neural network trained on natural images
  performs state-of-the-art image compression.
\newblock In {\em 2014 IEEE International Conference on Image Processing
  (ICIP)}, pages 4092--4096. IEEE.

\bibitem[Hopfield, 1982]{hopfield1982neural}
Hopfield, J.~J. (1982).
\newblock Neural networks and physical systems with emergent collective
  computational abilities.
\newblock {\em Proceedings of the national academy of sciences},
  79(8):2554--2558.

\bibitem[Hopfield et~al., 1983]{hopfield1983unlearning}
Hopfield, J.~J., Feinstein, D., and Palmer, R. (1983).
\newblock ‘unlearning’has a stabilizing effect in collective memories.
\newblock {\em Nature}, 304(5922):158--159.

\bibitem[Hsu, 2012]{hsu2012application}
Hsu, W.-Y. (2012).
\newblock Application of competitive hopfield neural network to brain-computer
  interface systems.
\newblock {\em International journal of neural systems}, 22(01):51--62.

\bibitem[Kanter and Sompolinsky, 1987]{kanter1987associative}
Kanter, I. and Sompolinsky, H. (1987).
\newblock Associative recall of memory without errors.
\newblock {\em Physical Review A}, 35(1):380.

\bibitem[Kepler and Abbott, 1988]{kepler1988domains}
Kepler, T.~B. and Abbott, L.~F. (1988).
\newblock Domains of attraction in neural networks.
\newblock {\em Journal de Physique}, 49(10):1657--1662.

\bibitem[Kleinfeld and Pendergraft, 1987]{kleinfeld1987unlearning}
Kleinfeld, D. and Pendergraft, D. (1987).
\newblock " unlearning" increases the storage capacity of content addressable
  memories.
\newblock {\em Biophysical journal}, 51(1):47--53.

\bibitem[Krauth and M{\'e}zard, 1987]{krauth1987learning}
Krauth, W. and M{\'e}zard, M. (1987).
\newblock Learning algorithms with optimal stability in neural networks.
\newblock {\em Journal of Physics A: Mathematical and General}, 20(11):L745.

\bibitem[Lansner, 2009]{lansner2009associative}
Lansner, A. (2009).
\newblock Associative memory models: from the cell-assembly theory to
  biophysically detailed cortex simulations.
\newblock {\em Trends in neurosciences}, 32(3):178--186.

\bibitem[Li et~al., 2016]{li2016modified}
Li, R., Qiao, J., and Li, W. (2016).
\newblock A modified hopfield neural network for solving tsp problem.
\newblock In {\em 2016 12th World Congress on Intelligent Control and
  Automation (WCICA)}, pages 1775--1780. IEEE.

\bibitem[Nadal et~al., 1986]{nadal1986networks}
Nadal, J., Toulouse, G., Changeux, J., and Dehaene, S. (1986).
\newblock Networks of formal neurons and memory palimpsests.
\newblock {\em EPL (Europhysics Letters)}, 1(10):535.

\bibitem[Pajares et~al., 2010]{pajares2010hopfield}
Pajares, G., Guijarro, M., and Ribeiro, A. (2010).
\newblock A hopfield neural network for combining classifiers applied to
  textured images.
\newblock {\em Neural Networks}, 23(1):144--153.

\bibitem[Parisi, 1986]{parisi1986asymmetric}
Parisi, G. (1986).
\newblock Asymmetric neural networks and the process of learning.
\newblock {\em Journal of Physics A: Mathematical and General}, 19(11):L675.

\bibitem[Pereira and Brunel, 2018]{pereira2018attractor}
Pereira, U. and Brunel, N. (2018).
\newblock Attractor dynamics in networks with learning rules inferred from in
  vivo data.
\newblock {\em Neuron}, 99(1):227--238.

\bibitem[Personnaz et~al., 1986]{personnaz1986collective}
Personnaz, L., Guyon, I., and Dreyfus, G. (1986).
\newblock Collective computational properties of neural networks: New learning
  mechanisms.
\newblock {\em Physical Review A}, 34(5):4217.

\bibitem[Recanatesi et~al., 2015]{recanatesi2015neural}
Recanatesi, S., Katkov, M., Romani, S., and Tsodyks, M. (2015).
\newblock Neural network model of memory retrieval.
\newblock {\em Frontiers in computational neuroscience}, 9:149.

\bibitem[Smith et~al., 1998]{smith1998neural}
Smith, K., Palaniswami, M., and Krishnamoorthy, M. (1998).
\newblock Neural techniques for combinatorial optimization with applications.
\newblock {\em IEEE Transactions on Neural Networks}, 9(6):1301--1318.

\bibitem[Storkey, 1997]{storkey1997increasing}
Storkey, A. (1997).
\newblock Increasing the capacity of a hopfield network without sacrificing
  functionality.
\newblock In {\em International Conference on Artificial Neural Networks},
  pages 451--456. Springer.

\bibitem[Storkey, 1999]{storkey1999cient}
Storkey, A.~J. (1999).
\newblock {\em E cient Covariance Matrix Methods for Bayesian Gaussian
  Processes and Hop eld Neural Networks}.
\newblock PhD thesis, Citeseer.

\bibitem[Taghizadeh-Sarabi et~al., 2013]{taghizadeh2013eeg}
Taghizadeh-Sarabi, M., Niksirat, K.~S., Khanmohammadi, S., and Nazari, M.
  (2013).
\newblock Eeg-based analysis of human driving performance in turning left and
  right using hopfield neural network.
\newblock {\em SpringerPlus}, 2(1):1--10.

\bibitem[Tanaka et~al., 2019]{tanaka2019spatially}
Tanaka, G., Nakane, R., Takeuchi, T., Yamane, T., Nakano, D., Katayama, Y., and
  Hirose, A. (2019).
\newblock Spatially arranged sparse recurrent neural networks for energy
  efficient associative memory.
\newblock {\em IEEE transactions on neural networks and learning systems}.

\bibitem[Tirdad and Sadeghian, 2010]{tirdad2010hopfield}
Tirdad, K. and Sadeghian, A. (2010).
\newblock Hopfield neural networks as pseudo random number generators.
\newblock In {\em 2010 Annual Meeting of the North American Fuzzy Information
  Processing Society}, pages 1--6. IEEE.

\bibitem[Van~Hemmen et~al., 1990]{van1990increasing}
Van~Hemmen, J., Ioffe, L., K{\"u}hn, R., and Vaas, M. (1990).
\newblock Increasing the efficiency of a neural network through unlearning.
\newblock {\em Physica A: Statistical Mechanics and its Applications},
  163(1):386--392.

\bibitem[Wen et~al., 2009]{wen2009review}
Wen, U.-P., Lan, K.-M., and Shih, H.-S. (2009).
\newblock A review of hopfield neural networks for solving mathematical
  programming problems.
\newblock {\em European Journal of Operational Research}, 198(3):675--687.

\end{thebibliography}



%







\end{document}